%% file: example.tex
\documentclass{article}
\usepackage{lipsum}
\usepackage[preprint]{corl_2026} 
\usepackage{graphicx}
\usepackage{amsmath}
\usepackage{amssymb}
\usepackage{tabularx}
\usepackage{hyperref}
\usepackage{wrapfig} 
\usepackage[T1]{fontenc}
\usepackage{xcolor}
\usepackage{algorithm}
\usepackage{algpseudocode}
\usepackage{booktabs} 
\usepackage[T1]{fontenc}
\usepackage{xcolor}
\newcommand{\pair}[2]{#1\hspace{0.2em}\rule[-0.2ex]{0.03em}{2ex}\hspace{0.2em}#2}

\title{\textsc{Mana}: Dexterous Manipulation of Articulated Tools}

\author{
  Zhao-Heng Yin$^{\text{1,4}}$\quad Guanya Shi$^{\dagger,\text{2,4}}$ \quad Pieter Abbeel$^{\dagger,\text{1,4}}$ \quad C. Karen Liu$^{\dagger,\text{3,4}}$\\
  UC Berkeley$^{\text{1}}$ \quad CMU$^{\text{2}}$ \quad Stanford University$^{\text{3}}$ \quad Amazon FAR$^{\text{4}}$\\ 
  \ \\
  Project Page: \href{https://zhaohengyin.github.io/mana}{\texttt{https://zhaohengyin.github.io/mana}}
}

\begin{document}
\maketitle
\begingroup
\renewcommand{\thefootnote}{}
\footnotetext{Correspondence to: Zhao-Heng Yin, \texttt{zhaohengyin@cs.berkeley.edu}. $\dagger$ indicates equal contribution.}
\endgroup
\vspace{-0.3cm}

\begin{abstract}
Articulated tool manipulation remains a major challenge in dexterous robotics due to the need to coordinate internal degrees of freedom and contact-rich interactions. While prior work has largely focused on rigid objects, articulated tool use remains underexplored because of its physical complexity and the difficulty of learning functional manipulation strategies. We present Mana (Manipulation Animator), a general sim-to-real framework that reinterprets dexterous manipulation as an animation problem. Inspired by computer animation, Mana employs a coarse-to-fine pipeline that transforms procedurally-generated grasp keyframes into manipulation trajectories through motion planning and reinforcement learning. The data generation process is largely automatic, requiring only a few mouse clicks to specify functional affordances (<1 minute per tool). Across four articulated tools spanning different scales and joint types, Mana achieves zero-shot sim-to-real transfer for both grasping and in-hand manipulation, demonstrating a scalable approach to dexterous articulated tool use.
\end{abstract}


\begin{figure}[h]
    \centering
    \includegraphics[width=1.0\linewidth]{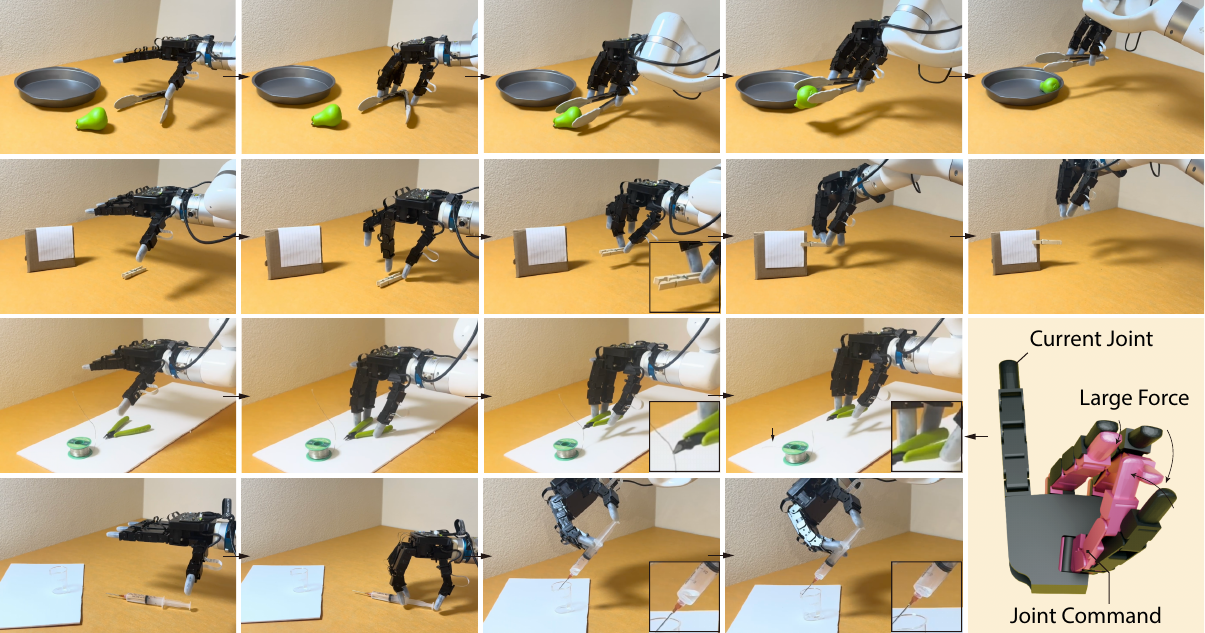}
    \caption{\textit{Mana}~(Manipulation Animator) is a framework for learning dexterous manipulation of articulated tools with zero-shot sim-to-real transfer. Our system can grasp and manipulate 4 types of tools of different challenging shapes, scales, and joint properties, including tongs, pliers, clothespins, and syringes. All grasping and finger control above are autonomous, except for the tool-to-site transition handled via wrist teleoperation. As shown in the bottom right, the seemingly simple process requires highly accurate force control to simultaneously stabilize and actuate the tool.}
    \label{fig:system_overview}
\end{figure}

\newpage
\begin{figure}[t]
    \centering
    \includegraphics[width=1.0\linewidth]{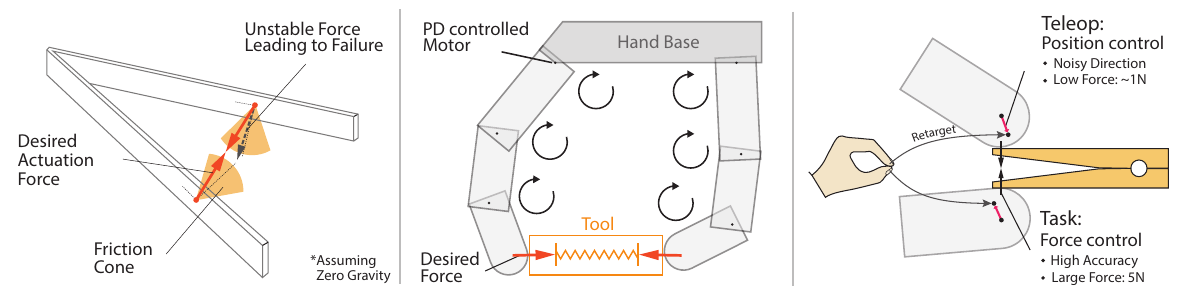}
    \caption{\textbf{Physical Challenges of Articulated Tool Use.} Left: Dexterous articulated tool manipulation is highly sensitive to contact points and force configuration. The fingers must apply precise contact force within the friction cone to actuate the tool stably. Middle: As the fingers and articulated tool form a tightly coupled dynamic system, even tiny execution errors at one joint can result in instability and failure. Right: Teleoperation is noisy and even infeasible. Existing teleoperation interfaces are position-based rather than force-based and fail to generate accurate or sufficient force.}
    \label{fig:challenge}
\end{figure}

\newpage
\section{Introduction}
Many of the most useful tools in human environments are not rigid objects, but articulated mechanisms. They require the user not only to grasp the tool, but also to actuate its internal degrees of freedom while maintaining stable contact with thin, moving handles. This capability is central to many practical manipulation tasks: picking up small objects with tongs, clipping a clothespin onto a surface, cutting a wire with pliers, or pushing a syringe plunger to dispense fluid. Despite rapid progress in robotic grasping and rigid-body dexterous manipulation~\cite{andrychowicz2020learning, lum2024dextrah,yin2025dexteritygen,lin2025sim,kedia2026simtoolreal}, autonomous grasping and manipulation of articulated tools with multi-fingered hands remains largely unsolved.

Articulated tool manipulation is difficult because the robot must simultaneously stabilize the tool and apply functional actuation forces. As illustrated in Fig.~\ref{fig:challenge}, These two objectives often require highly challenging force configurations within the friction cone. Forces that stably actuate the tool are not always aligned with the local surface normals, making the grasp prone to slipping. In contrast, forces aligned with the surface normals are less likely to slip, but can induce undesired accelerations or rotations of the tool body, ultimately destabilizing the grasp.  This tradeoff becomes even harder as multiple fingers and the articulated tool form a tightly coupled dynamic system constrained by complex hand kinematics. Small execution errors at one finger can change contact locations, alter moment arms, and destabilize the entire interaction. Finally, many tools require substantial force: cutting a wire with pliers or pushing a syringe plunger can require force magnitudes large enough to amplify both slippage and dynamic instability. Successful tool use therefore requires choosing contact locations and force directions that balance manipulation effectiveness, grasp stability, friction, moment arms, and hand kinematic feasibility as the tool configuration changes over time.

A natural way to obtain such behavior is through human teleoperation~\cite{handa2020dexpilot,sivakumar2022robotic, cheng2024open,ding2024bunny,yin2025geometric}, but articulated tool use exposes a key limitation of existing teleoperation interfaces (Fig.~\ref{fig:challenge}). Most dexterous teleoperation systems are primarily position-based: they retarget hand poses or fingertip positions rather than directly specifying contact forces. In practice, generating sufficient and well-directed pinch forces is difficult and even infeasible, especially for small, thin tools. A common workaround is to move the commanded fingertip target slightly inside the object along the local surface normal, causing the low-level PD controller to generate contact force. However, this method can generate only a limited force magnitude and is often insufficient to actuate stiff tool joints. In addition, the surface normal of an articulated tool may not coincide with the desired force direction, thereby limiting the effectiveness of the proposed workaround. On the other hand, sim-to-real reinforcement learning~\cite{andrychowicz2020learning, handa2023dextreme, yin2023rotating, chen2023sequential} approach is also insufficient on its own. Long-horizon articulated tool use requires the policy to discover precise functional contacts on centimeter-scale surfaces, maintain those contacts through changing tool configurations, and generate large task-specific forces in a high-dimensional hand action space. This makes exploration from scratch exceedingly challenging.

When real-world demonstrations are difficult to acquire and end-to-end reinforcement learning from scratch is too brittle, we turn to Computer Animation. However, unlike traditional animation pipelines, robotic manipulation data must be generated at scale, obey contact-rich physical constraints, and transfer reliably from simulation to the real world~\cite{liu2009dextrous,qin2022dexmv}. We introduce \textbf{Mana} (\textbf{Man}ipulation \textbf{A}nimator), an animation-inspired, scalable sim-to-real data generation framework, casting articulated tool use as a coarse-to-fine motion synthesis problem. It first uses a small amount of human input, with a few clicks specifying functional affordance regions on the tool mesh, to generate diverse grasp keyframes at scale. These keyframes define the skeleton of the manipulation sequence. Motion planning then connects collision-free reaching and pregrasp motions, while reinforcement learning is used only for the short, contact-rich phases where delicate position-force coordination is required. By decomposing long-horizon articulated tool manipulation into keyframes and short transition segments, Mana avoids the exploration difficulty of end-to-end RL while producing scalable simulation data for policy learning.

Once trajectories are generated, we return to machine learning for real-world deployment. We train a point-cloud-conditioned transformer diffusion policy that maps object observations to wrist and finger actions. The policy can perceive small articulated tools with a thickness of around 1 cm from RGB-D inputs, grasp them from a flat surface, and execute learned finger motions for functional tool actuation with 3-7N force. Importantly, all training data are generated in simulation, and the learned policies are transferred zero-shot to the real robot. We demonstrate the efficacy of our approach through a series of evaluations on four distinct articulated tools: tongs, pliers, syringes, and clothespins on an Allegro hand. These tools represent a broad spectrum of mechanical challenges, ranging from the compliant, long-range coordination of tongs to the high-impedance, precision-critical actuation of syringes and clothespins. Our results indicate that the proposed system achieves high success rates across these categories, exhibiting a new level of dexterity and sim-to-real transfer.

\begin{figure}[t]
    \centering
    \includegraphics[width=1.0\linewidth]{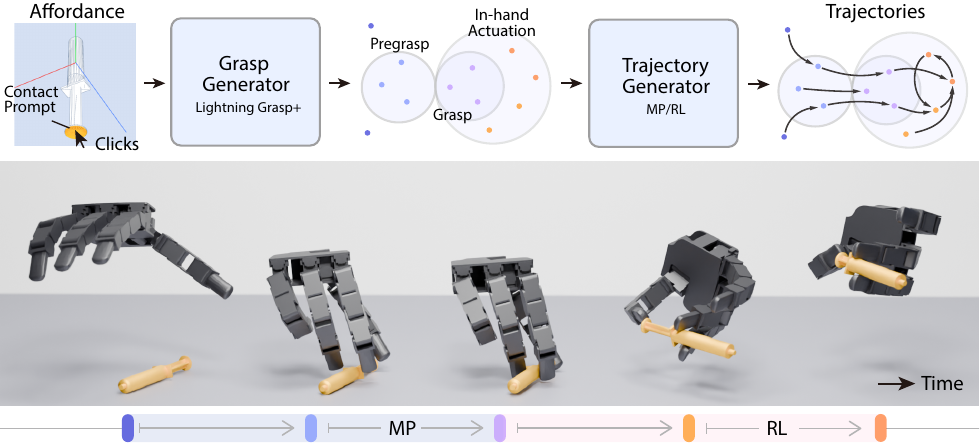}
    \caption{\textbf{Mana Data System Overview}. Mana takes a coarse-to-fine approach to generate tool manipulation data for policy learning. It decomposes the whole manipulation sequence with many \textit{procedurally} generated grasp \textit{keyframes}, and then use motion planning~(MP) and reinforcement learning (RL) to generate manipulation trajectories from these keyframes~(i.e., \textit{inbetweening}).}
    \label{fig:Mana_system}
\end{figure}
\section{Related Works}
Mastering dexterous object manipulation remains a fundamental challenge in robotics. Due to the high-dimensional configuration space and complex physics of multi-fingered hands, analytical and model-based methods are generally challenging or infeasible for manipulation, except for some specific cases involving regular geometries~\cite{pang2023global, chen2024springgrasp}. Consequently, research has shifted toward sim-to-real RL and imitation learning. While many works have demonstrated successful dexterous grasping of daily objects~\cite{shao2020unigrasp,wang2022dexgraspnet, qin2023dexpoint, wan2023unidexgrasp++, kannan2023deft, lum2024dextrah, singh2024dextrah, zhong2025dexgraspvla, lin2025sim, ye2025dex1b, chen2025dexonomy, rostel2025composing} and the in-hand reorientation or translation of rigid bodies using proprioception~\cite{qi2023general}, tactile sensing~\cite{yin2023rotating, khandate2023sampling, yang2024anyrotate, yuan2024robot, wang2024lessons, yin2025learning, hsieh2025learning}, and vision~\cite{andrychowicz2020learning, chen2023sequential, kedia2026simtoolreal}, dexterous manipulation of tools and articulated objects remains underexplored.

A seminal contribution in this domain is the Rubik’s Cube system~\cite{akkaya2019solving}, though it relied on object-specific designs. Other studies have investigated the manipulation of large articulated objects, such as doors, handles, laptops, boxes, and bottles, using either single-handed or bimanual setups~\cite{bao2023dexart,lin2024twisting,11059840,chen2024object,mandi2025dexmachina,li2025maniptrans,liang2026contrack}. 
In terms of tool use, some works focus on tools initialized in hanging or floating configurations that are relatively easy to perform a stable or power grasp~\cite{yin2024offline,wang2024dexcap,zheng2026egoscale} through imitation learning. Compared with these works, we tackle the more constrained and challenging problem of thin-tool precision grasping and using on tabletops using a demo-free, sim-to-real learning framework. Graphics research also studied more dexterous tool use such as chopsticks in simulation~\cite{yang2022learning}. However, how to transfer the learned policy into real world still remains an open problem. The most closely related works in robotics are by Xu et al.~\cite{11246691} and Atar et al.~\cite{atar2025hand}, which explored in-hand tool manipulation starting from a fixed, stable initial in-hand grasp through sim-to-real transfer. In contrast, our work addresses the entire pipeline for tiny and thin articulated tools, from initial tabletop object acquisition to functional manipulation, while accounting for the precise hybrid position-force control required during articulated tool use.

\section{Mana Data System}

Mana generates dexterous articulated-tool manipulation data through a coarse-to-fine pipeline. As illustrated in Fig.~\ref{fig:Mana_system}, the system first constructs a set of grasp keyframes that define the skeleton of the manipulation sequence, including pre-grasp states, stable grasp states, and in-hand actuation states. It then generates trajectories between these keyframes using the simplest mechanism appropriate for each phase: motion planning for geometric reaching motions and reinforcement learning for contact-rich transitions that require coordinated position-force control. This decomposition avoids the exploration burden of end-to-end RL while preserving the contact-rich behaviors needed for functional tool use. The input to Mana is a tool mesh with an articulated joint model and a small amount of human-provided functional affordance annotation. The output is a dataset of successful simulated manipulation trajectories, each containing wrist motion, hand joint commands, object state, and semantic phase labels such as grasping, opening, and closing. These trajectories are later used to train the visuomotor policy described in Sec.~\ref{sec:robot_policy}.

\subsection{Grasp Generator}
The grasp generator converts sparse functional affordance annotations into a dense set of physically plausible grasp and actuation states. For each tool, the user marks the canonical tool pose and the functional regions of the mesh in a 3D interface, such as the handles of pliers, the arms of tongs, or the plunger and barrel of a syringe. The labeling process is fast and takes less than 1 minute for each instance. We believe the process can be further automated by recent affordance models~\cite{yang2022oakink}. Then, we build upon the Lightning Grasp pipeline~\cite{yin2025lightning} to generate functional grasps with these annotations in the presence of environmental obstacles via a new collision-aware IK process~(Appendix A.1). Given the annotations, we query contact field to construct contact domains over the annotated surface. We then optimize fingertip contact locations within these domains and the finger joint positions. The output grasps are filtered for stability in IsaacLab simulation~\cite{mittal2025isaac}. 

Unlike grasp generation for rigid objects, articulated-tool manipulation requires not only a stable grasp, but a chain of grasps for future actuation. Therefore, the keyframe planner samples a dense set of grasp states across the functional contact regions and across relevant tool configurations. For example, for a syringe, it samples diverse states that support stable plunger actuation. This dense coverage of state space is important because articulated tool use is highly sensitive to millimeter-scale changes in contact location. The resulting keyframes define the nodes of the Mana trajectory graph, including pre-grasp, grasp, and in-hand actuation keyframes.

\subsection{Trajectory Generator}

The trajectory generator connects the keyframes produced by the planner into executable manipulation trajectories. Following the phase structure in Fig.~\ref{fig:Mana_system}, Mana decomposes each episode into pre-grasping, grasping, and in-hand tool actuation. This separation is important because each phase has different physical requirements and learning strategies. 

\paragraph{Pre-grasping} The pre-grasping phase moves the hand from an initial configuration to a collision-free approach pose near the target grasp. Since this phase does not require forceful interaction with the tool, reinforcement learning is unnecessary. We instead implement a GPU-accelerated RRT-Connect algorithm~\cite{kuffner2000rrt} with path smoothing to plan trajectories from randomly sampled initial hand configurations to synthesized pre-grasp poses in a scalable manner.  However, directly planning to the final grasp configuration often causes premature collision with the tool in the real world, which is especially problematic for thin articulated objects. We therefore define the pre-grasp pose by displacing the fingertips away from the final contact locations along the local surface normals by a randomly sampled distance, retaining only collision-free configurations. This gives the downstream grasping phase a consistent approach direction while preserving diversity in the generated dataset.

\paragraph{Grasping}
The grasping phase transitions from the pre-grasp pose to a stable tool grasp. In many cases, this transition can be generated procedurally. Given the optimized fingertip contacts, we compute a squeezing configuration by moving the fingertips inward along the contact normals. The hand is then commanded to close toward this configuration, secure the tool, and lift it from the table. Successful simulated trials are added to the training dataset. However, this geometric strategy is not always sufficient. For thin tools, unstable initial contacts, or geometries where contact normals do not align with the required stabilizing forces, grasping becomes a dynamic contact-rich control problem. In these cases, Mana trains a short-horizon RL policy to connect pre-grasp and grasp keyframes. We use the same RL formulation described in the next section for challenging cases.

\begin{figure}[t]
    \centering
    \includegraphics[width=1.0\linewidth]{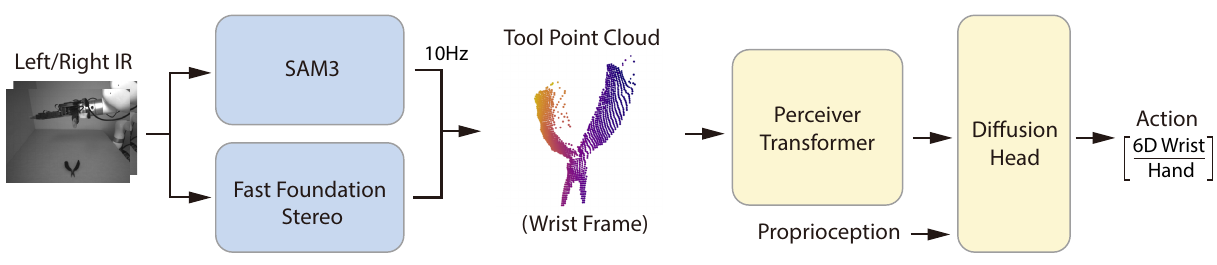}
    \caption{\textbf{Controller Architecture.} We use a point-cloud-based diffusion policy~(yellow modules) for control.  We train the policy with the successful manipulation trajectories generated by Mana.}
    \label{fig:controller}
\end{figure}

\paragraph{In-hand Tool Actuation} Once the tool is grasped, Mana generates in-hand actuation trajectories that change the internal configuration of the articulated tool while maintaining stable contact. These phases are inherently contact-rich: the hand must apply task-specific forces to the tool joint while preventing slip, ejection, or undesired motion of the tool body. We therefore use reinforcement learning to connect grasp and actuation keyframes. 

We frame in-hand tool actuation as a general goal-reaching problem. Each RL episode is initialized from a pre-generated stable grasp keyframe. The target is a randomly selected tool configuration corresponding to a meaningful tool-use transition. For example, if the tool is initialized in an open configuration, the target may be a corresponding closed configuration.  The reward combines three terms: $r = r_{\mathrm{tool}} + w_1r_{\mathrm{hand}} + w_2 r_{\mathrm{contact}}$. The first two terms $r_{\mathrm{tool}}$ and $r_{\mathrm{hand}}$ encourages tool and hand pose matching. The contact term $r_{\mathrm{contact}}$ encourages number of contact, and maintaining contact points to improve stability. Importantly, we use diverse force randomizations, such as random controller PD gain, object physical properties, random force perturbations on objects, and action noises, to train robust behavior. An episode terminates successfully when the tool reaches the target configuration and pose within a threshold. We leave the detailed reward and randomization design to the Appendix A.2. Successful trajectories are stored with phase semantics such as opening, closing, squeezing. These labeled trajectories form the simulation dataset used for policy learning.

\begin{figure}[t]
    \centering
    \includegraphics[width=1.0\linewidth]{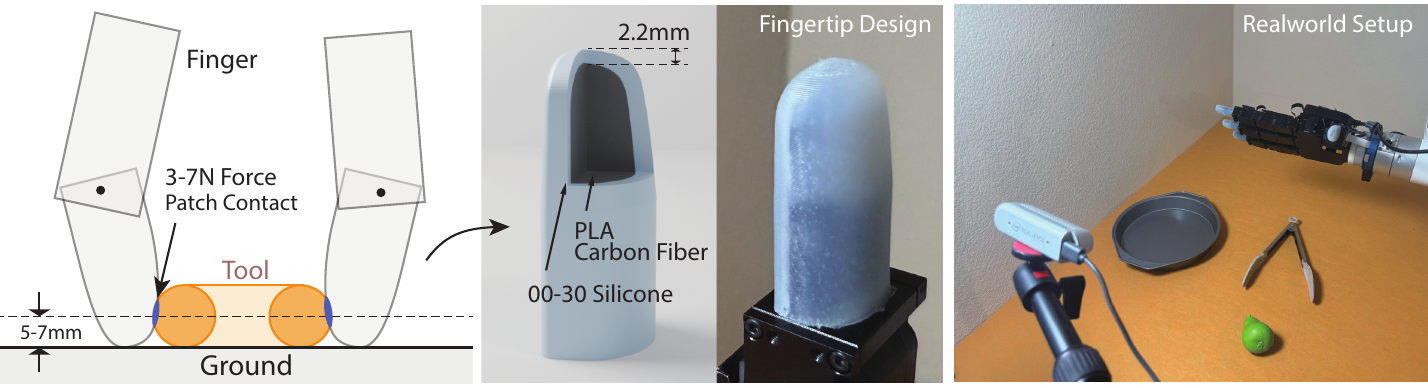}
    \caption{\textbf{Robot Hardware Setups}. Left: Fingertip Design. We find its shape and material critical for successful grasping from ground up and maintaining stable contacts during manipulation. Right: Deployment Environment Setup. We used a single Realsense D435 camera for perception.}
    \label{fig:hardware}
\end{figure}

\section{Robot System}

The Mana data system produces simulation trajectories, but real-world articulated tool use also depends critically on hardware design, perception, and policy execution. Our robot system consists of a dexterous manipulator platform and a point-cloud-conditioned diffusion policy trained entirely from Mana-generated simulation data.

\subsection{Dexterous Manipulator Platform}

Our platform uses a 7-DoF xArm7 robot arm equipped with a 16-DoF Allegro hand. Articulated tools impose unusually strict geometric and force requirements on the hand. Many of the tools in our experiments have thin handles or narrow contact surfaces, with object thicknesses on the order of 1 cm. Standard hemispherical rigid fingertips often create unstable point contacts on such geometries, leading to slip or tool ejection during forceful actuation.

To address this limitation, we design custom fingertips with a flattened, compliant contact surface. The fingertip profile increases the available contact patch when pressing against thin tool parts, while the soft silicone layer passively deforms around the tool surface. This compliance helps compensate for small pose errors and distributes normal forces more robustly during high-pressure actuation. 

For perception, the system uses an Intel RealSense D435 RGB-D camera. During deployment, we use SAM 3~\cite{carion2025sam} to segment the image, and the resulting mask is combined with Fast Foundation Stereo~\cite{wen2025fast} to produce an object point cloud. The point cloud is represented in the wrist frame before being passed to the policy. The current implementation runs at approximately 10 Hz on a workstation with two RTX 4090 GPUs.

\subsection{Visuomotor Policy for Articulated Tool Use}
\label{sec:robot_policy}

We train a point-cloud-conditioned diffusion policy to execute articulated tool manipulation from visual observations. The policy architecture is shown in Fig.~\ref{fig:controller}. Its input consists of the segmented tool point cloud in the wrist frame and robot proprioception. A Perceiver-style transformer~\cite{jaegle2021perceiver} first encodes the point cloud into a compact representation of a few tokens. These tokens, along with proprioceptive data, are then passed to a lightweight, transformer-based diffusion model~\cite{ho2020denoising} head for action generation. The policy outputs delta 6D wrist poses and delta hand joint position targets. The wrist is controlled by a differential IK solver, while the hand is managed by a low-level PD controller that generates the motor torque $\tau = K_p e + K_d \dot{e}$, 
where $e$ denotes the joint tracking error.

The policy is trained using successful Mana-generated trajectory dataset \(\mathcal{D}\). Let \(o_i\) denote the observation and \(a_i\) the corresponding action sampled from \(\mathcal{D}\). We train the diffusion policy with the standard sample-denoising objective:
\[
\mathcal{L}
=
\mathbb{E}_{(o_i,a_i)\sim \mathcal{D},\, t\sim U[0,T]}
\left[
\| a_i - \pi(\tilde{a}_i, o_i, t) \|^2
\right],
\]
where \(t\) is the diffusion timestep and \(\tilde{a}_i\) is the noisy action sample. To improve sim-to-real transfer, we apply point-cloud randomization during training, including noise perturbations and random part masking. This encourages the policy to tolerate imperfect segmentation, missing points, and depth noise during real-world deployment.

\begin{figure}[t]
    \centering
    \includegraphics[width=0.99\linewidth]{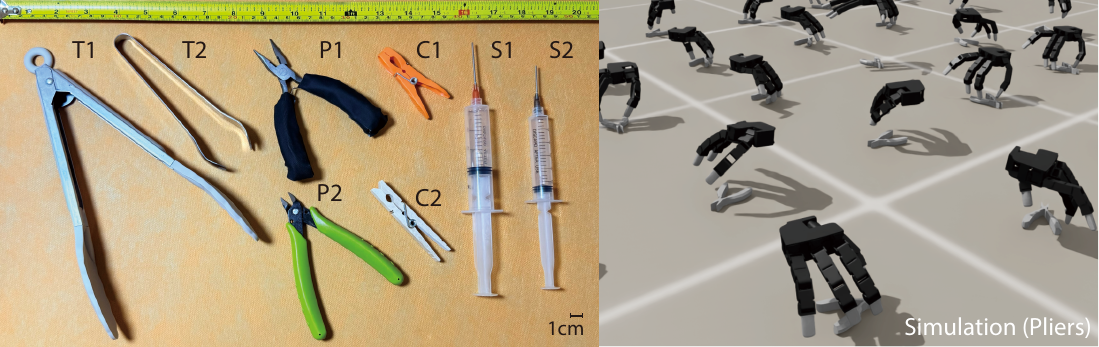}
    \caption{\textbf{Experimental Objects and Simulation Environment}. Our test objects cover different sizes, shapes, and joint properties. These tools have a thickness of \(\sim\)1cm and require 3-7N to actuate.}
    \label{fig:exp_objects}
\end{figure}

\section{Experiments}
\label{sec:experiment}
\subsection{Setups}
We evaluate our method on four categories of articulated objects: tongs, pliers, clothespins, and syringes. For each category, we use two object instances. As shown in Fig.~\ref{fig:exp_objects}, these objects cover a diverse range of sizes, shapes, and joint types.The objects have thicknesses of approximately 0.8–1.5cm, which is relatively small relative to the Allegro hand used in our experiments (approximately 2\(\times \) the size of a human hand). This makes grasping and manipulation particularly challenging, as actuation requires 3–7 N. We generate meshes and URDF models using camera-based scanning and feed them into the Mana pipeline for training the point-cloud-based policies. We compare our method against two baselines: open-loop Mana policy and teleoperation. For the open-loop baseline, we manually initialize the object pose and execute the corresponding Mana-generated trajectory. For the teleoperation baseline, a human operator controls the robotic hand using the Geometric Retargeting~(GeoRT)~\cite{yin2025geometric} system to manipulate the objects. The operator is given 1 hour to practice with the objects before evaluation. We evaluate different phases separately and ensure that the policy begins at a stable initial pose for that phase.

\subsection{Main Results}
As shown in Table~\ref{tbl:results}, our system successfully manipulates objects across all categories, achieving approximately 70\% success rates for both grasping and in-hand manipulation. In contrast, both baseline methods struggle in these scenarios. As the used object mesh is not perfect and pose estimation is subject to small perception errors, openloop finger motions can introduce contact errors or perturbations before finger closure, leading to grasp failures or unsuccessful manipulation. Teleoperation shows limited performance due to the difficulty of precisely controlling the force when manipulating thin articulated objects. It can only achieve around 30\% success on tongs, echoing recent findings~\cite{wang2026dexjoco}. The small object thickness and complex contact dynamics make reliable grasping and in-hand manipulation challenging even with extensive training.  We further observe frequent tool slippage and bouncing caused by imbalanced contact forces during teleoperation. Besides, the position-based retargeting method fails to generate sufficient contact force to actuate the clothespin.

\subsection{Ablation Studies}
\begin{figure}[t]
    \centering
    \includegraphics[width=1.0\linewidth]{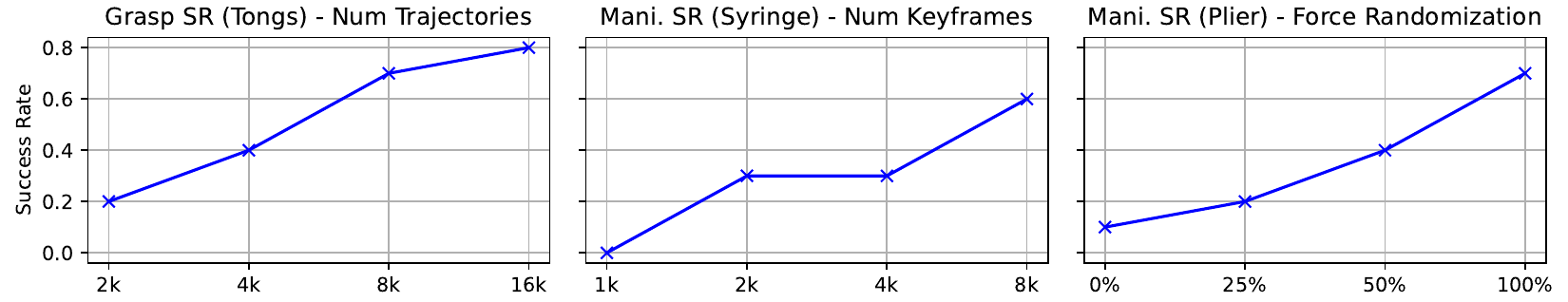}
    \caption{\textbf{Ablation Study}. Task success strongly depends on data quantity, state diversity, and force randomization. Since the desired force direction and magnitude are highly dependent on the contact points, increasing these factors enhances robustness in our contact-sensitive manipulation tasks.}
    \label{fig:ablation}
\end{figure}

In this section, we look into the importance of our data design. We evaluate the performance of our trained policy using different level of data diversity. As shown in Figure~\ref{fig:ablation}, we find that the real world performance scales with the number of trajectories used and the number of grasp keyframes used. The result suggests that for dexterous articulated tool manipulation, the robustness scales with the state coverage in simulation, which echoes with recent findings~\cite{yin2026emergent}. Tool manipulation relies on highly delicate precision grasps, where even millimeter-level discrepancies at the contact point can drastically alter force behavior. Consequently, densely sampling grasp configurations around functional poses to explore diverse contact modes is essential for learning stable multi-point position-force control. Furthermore, the robustness of data collection heavily depends on the intensity of force perturbations and action randomization. In real-world deployment, robotic actuators frequently experience noise or torque degradation due to overheating. To ensure controllers remain resilient against these and maintain a stable balancing force, integrating sufficient object and action perturbations during training is crucial.
\begin{table}[t]
    \centering
    \resizebox{\linewidth}{!}{
        \begin{tabular}{l *{11}{c}}
            \toprule
            & \multicolumn{3}{c}{Tongs} & \multicolumn{3}{c}{Pliers} & \multicolumn{3}{c}{Clothespins} & \multicolumn{2}{c}{Syringes} \\
            \cmidrule(lr){2-4} \cmidrule(lr){5-7} \cmidrule(lr){8-10} \cmidrule(lr){11-12}
            Method & Grasp & Open & Close & Grasp & Open & Close & Grasp & Open & Close & Grasp & Use \\
            \midrule
            Teleop (GeoRT) & \pair{0.3}{0.3}  & \pair{0.1}{0.2} & \pair{0.3}{0.3} & \pair{0.2}{0.1} & \pair{0.1}{0.1} & \pair{0.1}{0.0} & \pair{0.0}{0.0} & \pair{0.0}{0.0} & \pair{0.0}{0.0} & \pair{0.0}{0.0}  & \pair{0.0}{0.0}  \\
            Ours (Openloop)  & \pair{0.5}{0.6}  & \pair{0.6}{0.7} & \pair{0.4}{0.6} & \pair{0.4}{0.4} & \pair{0.3}{0.4} & \pair{0.3}{0.3} & \pair{0.3}{0.2} & \pair{0.2}{0.4} & \pair{0.4}{0.3} & \pair{0.1}{0.0}  & \pair{0.3}{0.3}  \\
            Ours     &  \pair{\textbf{0.8}}{\textbf{0.8}} & \pair{\textbf{0.8}}{\textbf{0.8}} & \pair{\textbf{0.7}}{\textbf{0.8}} & \pair{\textbf{0.7}}{\textbf{0.6}} & \pair{\textbf{0.7}}\textbf{{0.7}} & \pair{\textbf{0.7}}{\textbf{0.7}} & \pair{\textbf{0.8}}{\textbf{0.7}} & \pair{\textbf{0.8}}{\textbf{0.7}} & \pair{\textbf{0.6}}{\textbf{0.7}} & \pair{\textbf{0.7}}{\textbf{0.5}} & \pair{\textbf{0.6}}{\textbf{0.6}} \\
            \bottomrule
        \end{tabular}
    }
    \caption{\textbf{Experimental Results}. Our system outperforms baseline approaches across different phases and objects. Each evaluation contains 10 trials. The left and right numbers in each cell indicate the success rates of instances 1 and 2, respectively.}
    \label{tbl:results}
\end{table}

\subsection{Demonstrating Articulated Tool Use} 
\begin{wraptable}{r}{0.3\textwidth}
    \vspace{-22pt} 
    \centering
    \caption{Success rates (SR) across diverse tool use tasks.}
    \vspace{3pt}
    \label{tab:tool_use_results}
    \begin{tabularx}{\linewidth}{lc}
        \toprule
        \textbf{Task} & \textbf{SR} \\
        \midrule
        Tong Pick       & 7/10   \\
        Plier Cut       & 5/10   \\
        Clothespin Use  & 6/10   \\
        Syringe Inject  & 5/10   \\
        \bottomrule
    \end{tabularx}
    \vspace{-10pt}
\end{wraptable}
Having acquired several deployable in-hand manipulation skills, we can compose them into complex, functional tasks. However, precisely aligning a tool with a target object (e.g., 0.5mm thin wires) remains challenging due to perception limitations. To isolate this issue, we employ manual wrist teleoperation exclusively for the fine-alignment phase. We evaluate our policy's ability to successfully execute the overall tool-use workflows. In these trials, wrist teleoperation is enabled only after the object has been successfully grasped and lifted by the policy. Quantitative evaluation results across these tasks are summarized in Table~\ref{tab:tool_use_results}. As the tools interact with other objects, they experience unseen perturbations and may slip into non-recoverable poses, which degrades the performance. We hypothesize that better force and contact physics modeling in simulation can help to address these issues.

\section{Limitations}
We identify the following limitations in this work. First, due to insufficient motor strength (maximum torque of 0.7 Nm and a longest finger link of 0.06–0.07 m), our system cannot handle common stiff tool-use cases where the required force or activation threshold exceeds 10 N (e.g., trigger mechanisms). In addition, we focus on precision grasps and do not explore power-grasp-based tool use strategies as humans do, since the Allegro hand is significantly larger (2× human hand size), making it difficult to hold and actuate thin handles designed for human hands using power grasp. We believe these limitations can be addressed in future work with improved hardware design. On the perception side, detecting and responding to slip under occlusion, as well as manipulating very small tools, remains challenging. These tasks require high-frequency tactile and force sensing to resolve subtle changes in tangential friction at the fingertips. Finally, we used wrist teleoperation to combine different phases to test the feasibility of skill chaining, and we need additional policy to make the skill execution fully autonomous in the future.


\acknowledgments{The authors thank Tingfan Wu for assistance with hardware-related inquiries.}


\bibliography{example}  
\include{appendix_text}

\end{document}

%% file: appendix_text.tex
\appendix
\begin{figure}
    \centering
    \includegraphics[width=\linewidth]{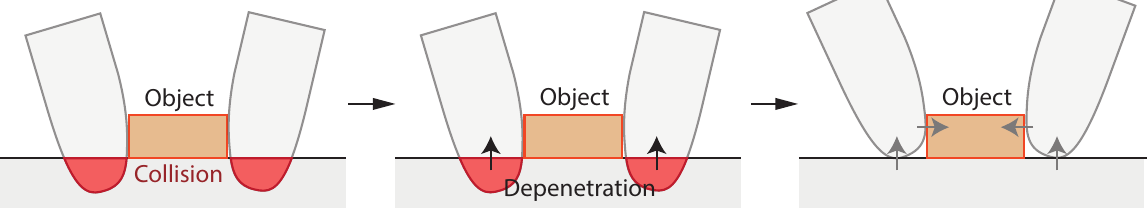}
    \caption{\textbf{Collision-aware Kinematics Optimization Procedure in Lightning Grasp+~(LG+).} When collisions occur during optimization~(Left), we will generate depenetration finger movements and add them to IK objectives~(Middle). This will resolve the collision while maintaining contacts~(Right). The procedure is essential for generating grasps of thin objects resting on the ground.}
    \label{fig:elg}
\end{figure}
\begin{figure}
    \centering
    \includegraphics[width=1\linewidth]{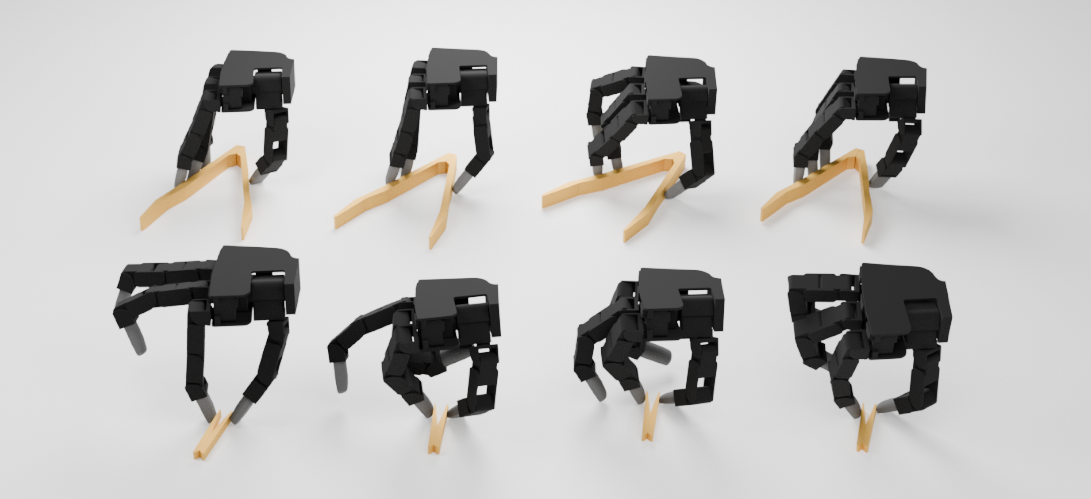}
    \caption{\textbf{Tabletop Grasp Samples Produced by LG+ System.} With the depenetration procedure in Fig.~\ref{fig:elg}, we can generate diverse challenging tabletop grasps. The thickness of these objects are around 1cm. We find these grasps difficult for RL to discover through direct exploration. Starting from these poses, learning to grasp using either RL or MP becomes significantly simpler.}
    \label{fig:grasps}
    \vspace{-0.3cm}
\end{figure}

\section{Implementation}
\subsection{Enhanced Lightning Grasp}
A limitation of the Lightning Grasp system is its inability to handle narrow grasping surfaces close to the ground or object corners~(Fig.~\ref{fig:elg}). This scenario is common for flat objects resting on the ground. Its kinematic optimization procedure is not collision-aware, and the resulting finger placement will directly collide with the neighborhood surface. The fingertip placement has to be very precise to maintain object contact while avoiding ground collisions. 

To address this issue, we propose a collision-aware kinematics optimization procedure. The resulting system is called Lightning Grasp+. The idea is to exert a virtual force to pull the fingertip away from the collided body in each iteration, while approaching the latest identified contact point. In this way, we can synthesize grasps that original Lightning Grasp fails to handle. More specifically, for each finger link colliding with the object, we detect the penetration center $p$ and the averaged separation vector $d$. During each kinematic optimization step, the joint configuration $q$ is updated by optimizing two objectives (Right of Fig.~\ref{fig:elg}): (1) resolving penetrations by pulling the finger away from surrounding objects through matching $p$ to $p+d$ when penetration occurs, and (2) dragging the finger surface point toward the identified contact point, following the LG system. The whole procedure is shown in Algorithm~\ref{alg:ik}. The procedure is implemented with CUDA for efficient parallel computation. 

\begin{algorithm}[h]
\caption{Collision-aware Kinematics Optimization}
\label{alg:ik}
\begin{algorithmic}[1]
\State \textbf{Input:} Initial Hand Joint $q_0$, Initial Target Hand Contact $x_0$, Object Contact $c$, Object Mesh $O$, Hand Model $H$. Total Optimization Steps $n$;
\State \textbf{Output:} Optimized Hand Joint $q$ and Contact $x$.

\State $q\leftarrow q_0, x\leftarrow x_0;$
\For{$i \in 1,2,...,n$}
    \State $x\leftarrow {\rm NearestProjection}(q, x, c, O, H)$;
    \State $p,d\leftarrow {\rm Penetrations}(q,O,H)$; ($p$ is center, $d$ is separation vector)
    \State $q\leftarrow {\rm LMStepIK}(x\to c; p\to p+d; q, O,H)$; (Levenberg–Marquardt IK method)
\EndFor

\State \textbf{Return:} $q, x$;
\end{algorithmic}
\end{algorithm}

\subsection{RL: Task Reward Design}
As discussed in the main text, we use the following dense reward function
\[
r = r_{\text{tool}} + w_1 r_{\text{hand}} + w_2 r_{\text{contact}}
\]
to train our RL policies. The detailed definitions of each reward component are provided below. Most terms follow an exponential formulation, consistent with standard practices in motion tracking.

\paragraph{Tool Reward} The tool reward is defined as
\begin{equation}
    r_{tool} = \exp (-\beta_{11} \Vert \mathbf{q}-\tilde{\mathbf{q}}\Vert_1) + \lambda_1\exp(-\beta_{12}  \Vert \mathbf{x}-\tilde{\mathbf{x}}\Vert_2-\beta_{13} \Vert\log(\mathbf{R}^{\mathsf{T}}\tilde{\mathbf{R}})^\vee\Vert_2).
\end{equation}
Here, $\mathbf{q}$ and $\tilde{\mathbf{q}}$ denote the current and target tool joint positions, respectively. $\mathbf{x}$ and $\tilde{\mathbf{x}}$ represent the current and target tool positions in the hand frame, while $\mathbf{R}$ and $\tilde{\mathbf{R}}$ denote the corresponding current and target tool orientations in the same frame. Note that for $X \in SO(3)$, $\log(X)^\vee$ denotes the axis-angle representation of $X$, whose norm corresponds to the rotation angle. Consequently, $\|\log(\mathbf{R}^{\mathsf{T}}\tilde{\mathbf{R}})^\vee\|_2$ defines the geodesic rotation distance between the current and target orientations.

\paragraph{Hand Reward} The hand reward for tool use is defined as:
\begin{equation}
    r_{hand} = \exp (-\beta_{21} \Vert \mathbf{q}_h-\tilde{\mathbf{q}}_h\Vert_1).
\end{equation}
Here, $\mathbf{q}_h$ and $\tilde{\mathbf{q}}_h$ denote the current and target hand joint positions. This reward is used for regularizing the hand gesture.

For the grasping case, the hand reward further incorporates a general wrist matching term.
\begin{equation}
    r_{hand} = \exp (-\beta_{21} \Vert \mathbf{q}_h-\tilde{\mathbf{q}}_h\Vert_1) + \lambda_2 \exp(-\beta_{22}  \Vert \mathbf{x}_h-\tilde{\mathbf{x}}_h\Vert_2-\beta_{23} \Vert\log(\mathbf{R}^{\mathsf{T}}\tilde{\mathbf{R}})^\vee\Vert_2).
\end{equation}
Here, $\mathbf{x}$ and $\tilde{\mathbf{x}}$ represent the current and target palm positions in the world frame, while $\mathbf{R}$ and $\tilde{\mathbf{R}}$ denote the corresponding current and target palm orientations in the same frame. 
 
\paragraph{Contact Reward} The contact term is simply counting the number of active contacts. 
\begin{equation}
    r_{contact} = \sum_{i\in \text{Finger},j\in\text{Tool}} \mathbf{1}[f_{ij} > \epsilon].
\end{equation}
Here, $f_{ij}$ denotes the contact force between finger link $i$ and tool link $j$. The force reading is obtained from IsaacLab contact sensor.

The setup of these hyperparameters are as follows. $w_1=0.1, w_2=0.15, \lambda_1=0.3, \lambda_2=10.0, \beta_{12}=100.0, \beta_{13}=5.0, \beta_{21}=10.0,\beta_{22}=50.0,\beta_{23}=5.0$. Due to difference in units, for revolute joint, we use $\beta_{11}=10.0$; for prismatic joint~(syringe), we use $\beta_{11}=50.0$. 

Besides these, we also use +100 bonus when the goal is achieved. We also reset the environment with a -50 penalty when the object falls down to the ground or deviate too much from the target grasp. Finally, the total reward is scaled by 0.1 during PPO training to improve optimization stability.

\subsection{RL: Domain Randomizations}
We find it essential to include diverse force-related domain randomizations to improve the physical robustness of (teacher) RL policies. Specifically, we include the following:
\begin{itemize}
    \item Action Noise: $+U[-0.1,0.1]$. This is directly added to the input action inside each environment step. This simulates motor noise during execution. The original action scale is from $-1$ to $1$. Therefore, this corresponds to approximately $10\%$ noise.

    \item Robot PD Controller Gain: $\times U[0.8, 1.2]$. This randomly scales the robot’s low-level PD gains to model variability in actuator stiffness and control tuning. It captures uncertainty in the robot’s internal control dynamics across deployments.

    \item Tool PD Controller Gain: $\times U[0.8, 1.5]$. This perturbs the PD gains associated with tool-level stabilization and actuation, modeling differences in tool dynamics such as compliance, friction, and attachment variability.

    \item Mass: $\times U[0.8, 1.2]$. This ensures policy robustness across different tool masses.
    \item Friction: $\times U[0.5, 1.5]$. This randomizes the contact friction coefficient to prevent overfitting to a single friction configuration and encourages robustness across a wide range of surface interaction conditions.
    
    \item Object Force Perturbation: We randomly apply Gaussian force perturbations to the object at each step. Each force is sampled from $\mathcal{N}(0, \sigma^2)$, where $\sigma = \alpha m g$, and $\alpha \sim U[0.05, 0.2]$, with $m$ the object mass and $g$ the gravitational constant. Each perturbation persists for several consecutive steps before resampling.
\end{itemize}
\subsection{RL: Training}
We train the RL policy in IsaacLab using the PPO algorithm. Both the actor and critic are parameterized as multilayer perceptrons (MLPs) with four hidden layers of size $[512, 512, 512, 512]$. They take privileged state information as input, including hand and tool poses and configurations, along with their velocities, contact forces, goal poses and configurations, and control targets. The observations are normalized with an online observation statistics tracker during learning.

For PPO training, we use a discount factor $\gamma = 0.99$, clipping $\epsilon=0.2$, GAE parameter $\lambda = 0.95$, $4096$ parallel environments, and a minibatch size of $8192$. We perform PPO updates every 8 environment steps, running 5 optimization epochs on the collected rollout data. The initial learning rate is set to $1\times10^{-4}$, and is adjusted online using a KL-divergence-based learning rate scheduler with threshold KL 0.01.

\subsection{Visuomotor Policy: Architecture}
Our (student) visuomotor policy takes point-cloud-based tool observations and proprioception as input, and outputs delta wrist pose and hand joint positions. We represent the tool using a point cloud of 512 points in the wrist frame; when the number of points differs, we first resample to obtain a fixed-size representation. The point cloud is then encoded using a Perceiver Transformer. Specifically, we use 4 learnable query tokens. Each block consists of a self-attention layer over the queries, a cross-attention layer between the queries and the point cloud, and a feedforward MLP. We use 4 attention heads with a latent dimension of 128. This encoder will output four 128-dimenstional tokens. The proprioception input consists of the hand joint positions and their corresponding targets from past two frames. It is encoded using an MLP with hidden dimensions $[512, 256, 256]$ and projected into a 128-dimensional token.

We then employ a transformer-based diffusion head to generate actions, conditioning on the encoded 128-D tokens. We find that predicting a single future action is sufficient for control. The diffusion head consists of 6 layers and 2 attention heads, with a latent dimension of 256. 

We train the policy using the AdamW optimizer with a learning rate of $1\times10^{-4}$, a batch size of 128, and a weight decay of 0.01. We also incorporate a cosine learning rate scheduler with linear warmup of 1000 steps. The learning rate is decayed to $1\times 10^{-6}$ at $2\times 10^5$ training steps. During control, we use an exponential moving average (EMA) of the model parameters.

\subsection{Force Simulation}
We find acurate calibration of force-related parameters in simulation to be critical for successful sim-to-real transfer. To this end, we perform system identification on the robot hand to ensure that its force-response characteristics closely match those observed on the real hardware.  

For the tools, we approximate their force response using a spring--mass--damper model, since most tools considered in our experiments are driven by internal springs. We estimate the model parameters using measurements collected with a electronic force gauge and implement the resulting behavior through a PD controller with internal joint frictions in simulation. Although the identified PD gains are only approximate, the aggressive randomization of force-related parameters during training reduces sensitivity to modeling errors and improves robustness during real-world deployment.

As the stiffness of the tool and the interaction force are quite high in our experiments, we find it crucial to use Implicit Euler method for stable actuator simulation as opposed to Explicit Euler as in previous works. Furthermore, large forces on 0.5-1cm scale tiny objects can easily lead to penetrations. We also use significantly larger position~(16-32) and velocity iterations~(4-6), with a smaller $dt$~(1/200s) in IsaacLab to ensure stability.

\subsection{Fingertip Shape Design and Fabrication}
We generate the fingertip geometry using the following procedure. First, we define a canonical cross-sectional boundary $C_0$ on the plane $z=0$, represented as a closed curve in the $XY$ plane. This boundary is then extruded along the positive $z$-axis with anisotropic scaling. Specifically, the cross-section at height $z=h$ is defined as

\[
C_h = \left\{ \bigl(x\,w_x(h),\, y\,w_y(h),\, h\bigr)\;\middle|\; (x,y,0)\in C_0 \right\},
\]

where $w_x(h)$ and $w_y(h)$ denote the scaling factors along the $x$- and $y$-axes, respectively. The canonical boundary $C_0$ and the scaling functions $w_x(\cdot)$ and $w_y(\cdot)$ are designed to approximate the cross-sectional profile of a human fingertip. Both scaling functions monotonically decrease to zero at the fingertip apex at $z=H$, i.e., $w_x(H)=w_y(H)=0$.

The final fingertip geometry is defined as the convex hull of the union of all cross-sections:

\[
M = \operatorname{ConvHull}\!\left( \bigcup_{z\in[0,H]} C_z \right).
\]

This construction yields a smooth, tapered shape that closely resembles the geometry of a human fingertip. We generate the fingertip skeleton by computing an offset mesh of the fingertip geometry. Both the mold and the skeleton are fabricated using 3D printing. The skeleton is then secured inside the mold, and silicone is injected into the cavity between the two components. After the silicone cures, the molded fingertip is removed from the mold and ready for use.